\documentclass[conference]{IEEEtran}
\IEEEoverridecommandlockouts
\usepackage{booktabs}
\usepackage{array}
\usepackage{multirow}
\usepackage{amssymb}
\usepackage{url}
\usepackage{titlesec}
\usepackage{tikz}
\usepackage{tabularx}
\usetikzlibrary{positioning, arrows.meta, shapes.geometric, fit, calc}

\providecommand{\doi}[1]{doi:\,#1}

\titlespacing*{\section}{1pt}{1pt}{0.5pt}
\titlespacing*{\subsection}{0pt}{0.5pt}{0.25pt}

\begin{document}



\title{Natural-Language Zero-Knowledge Verification via Skeptical MCP Tools}

\author{\IEEEauthorblockN{Md Nazrul Huda Shanto \quad Paritosh Ramanan}
\IEEEauthorblockA{Oklahoma State University, Stillwater, OK, USA\\
\{mshanto, paritosh.ramanan\}@okstate.edu}}


\maketitle
{\renewcommand{\thefootnote}{}\footnotetext{\copyright~2026 IEEE. Personal use of this material is permitted. Permission from IEEE must be obtained for all other uses, in any current or future media, including reprinting/republishing this material for advertising or promotional purposes, creating new collective works, for resale or redistribution to servers or lists, or reuse of any copyrighted component of this work in other works. Accepted for publication at the 2026 IEEE International Conference on Big Data (IEEE BigData 2026).}}
\begin{abstract}

Zero-knowledge proofs enable verifying computation without revealing the underlying datasets. Recent studies have demonstrated the practicality of such frameworks for critical application domains such as healthcare. However, analyzing large scale chained proof artifacts introduces challenges in interpretability and usability barriers for non-expert stakeholders which can potentially be resolved using LLM based tools. On the other hand, using MCP tools with LLM integrations introduces a new risk pertaining to correctness of parameters supplied during the proof generation and verification lifecycle. In this paper, we present a natural-language interface to zero-knowledge verifiable computation that is based on the Model Context Protocol (MCP). We introduce a \textbf{skeptical tools} approach that does not place implicit trust in model inputs, but leverages server-side state estimates for checks, validations and completions pertaining to user queries. Using large-scale experimental evaluations across three open-weight model families involving 6000 queries, we demonstrate that our skeptical approach raises task completion rates by up to 23\% over conventional methods settling at 93\% across for Qwen family of models. Our results ultimately yield concrete, quantified insights into the occurrence of correctness failures, fabrication, and privacy exposures at the interface between LLM agents and zero-knowledge proof systems.
\end{abstract}


\section{Introduction}

Zero-knowledge proof frameworks enable verifiable computation over private datasets without public disclosure, and have recently been applied to healthcare settings including clinical trial analyses~\cite{shahid2026cosmetic} and genomic studies~\cite{frolov2026icefish,zhang2025toward}. To balance verifiability, privacy, and tractability, these frameworks decompose the underlying computation into sequential subgraphs, each producing an independent proof artifact~\cite{shahid2026cosmetic,frolov2026icefish}. As user bases grow, the resulting system of interlinked artifacts scales linearly, and asserting their integrity and mutual consistency becomes a non-trivial verification burden. Such complexities create a critical usability gap, limiting the ability of auditors and regulators to interpret verification outcomes efficiently~\cite{scaramuzza2025zkaudit,carreira2021usable}. The rise of agentic tool-calling in large language models (LLMs) offers an opportunity to bridge this gap through a natural language interpretive layer that abstracts verification functions across interdependent proof chains.

We ground our investigation in the CoSMeTIC framework~\cite{shahid2026cosmetic}, which applies zkSNARKs~\cite{groth2016snark} to clinical statistical analyses including Kolmogorov-Smirnov (KS), logistic regression accuracy (ACC) and ratio (LRT) tests. CoSMeTIC produces a system of multi-circuit zkSNARK artifacts that must remain consistent across users and computational subgraphs. While it avoids disclosing patient records, its reliance on programmatic API calls with precise hash values and nonce parameters creates a significant usability barrier for non-expert stakeholders such as trial participants and regulators.

A conversational LLM interface (e.g., \emph{``prove my data has been used in the Kolmogorov-Smirnov test''}) is a natural solution to the usability problem. However, integrating LLMs into cryptographic pipelines introduces serious reliability risks. Enhanced reasoning can causally increase tool hallucination~\cite{yin2025reasoning}, errors cascade as models justify incorrect answers with further false claims~\cite{zhang2024snowball}, and agents report failures honestly only when explicitly signaled, with dishonesty reaching 45\% in unsignaled cases~\cite{sethi2026fabrication}. Architectural design, moreover, outweighs model size in determining overall reliability~\cite{toda2025chase}.

In the zero knowledge context these risks carry elevated consequences. An incorrect proof-type selection generates a cryptographically valid proof for the wrong statistical test, one that passes standard verification while masking semantic intent. Fabricated or leaked user identifiers bind proofs to incorrect datasets, with direct regulatory implications in healthcare. Since the LLM acts as an intermediary, sensitive artifacts such as dataset hashes and job identifiers traverse a boundary outside the proof system's own privacy guarantees.

We therefore ask: \emph{can an LLM agent reliably operate a chained, multi-circuit zkSNARK system?} We note that reliable operation requires the agent to accurately select analyses, preserve user-dataset binding, maintain proof lineage across stateless requests, and report status without hallucination. Each step is a potential failure point when LLM-mediated.

To investigate this, we develop a containerized LangGraph-based ReAct agent~\cite{yao2023react} integrated with CoSMeTIC via a FastMCP server exposing eight proof-related tools through MCP~\cite{hou2025mcp}. We conduct evaluations across three state-of-the-art LLMs (Qwen-3 32B, Mistral Small 3.2 24B, and Llama-3 8B) using natural language queries (e.g., ``prove that my data is in KS'') to trigger specific tool calls. Our design employs a \textbf{skeptical tools} pattern wherein MCP tools verify, complete, and override LLM-generated parameters at execution time to ensure cryptographic correctness.

To the best of our knowledge, this work presents the first evaluated natural-language interface for zero-knowledge verifiable computation of statistical analyses, and the first ablation study of server-side corrections for MCP agent tool-call parameters. Our specific contributions are as follows.

\begin{itemize}
\item  We demonstrate that zkSNARK-based verifiable computation is reliably accessible via natural language. Our full system achieves a 93\% task completion on Qwen-3 32B and completes the entire proof lifecycle in every run.

\item We propose a novel architectural pattern where MCP tools actively verify and correct LLM outputs, comprising 20 labeled decisions and two toggled behaviors, significantly enhancing robustness against LLM imprecision and hallucination.

\item  We identify critical failure modes where in the absence of proof-types incorrect provers are invoked, producing inaccurate user replies. We propose server-side correction mechanisms to eliminate this error across all tested model configurations.

\item  We present an 82-query ablation study across five configurations, a corrected faithfulness grader, and a threat model mapping each cryptographic asset to the LLM behaviors that endanger it, including metadata exposure to model providers.
\end{itemize}

\section{Related Work}
Zero-knowledge proofs (ZKP) allow a prover to convince a verifier of a statement's truth without revealing supporting information~\cite{goldwasser1989knowledge}. zkSNARKs extended this paradigm with succinct non-interactive proofs~\cite{groth2016snark,parno2013pinocchio} where commitment schemes help bind private inputs to the proofs, making identity and uniqueness critical for end-to-end verification. However, two cryptographic properties expose a persistent gap between mathematical soundness and semantic intent. Soundness guarantees that a proof verifies only if the encoded circuit holds, but not that the circuit matches the user's intended computation~\cite{shahid2026cosmetic}. As Chaliasos et al.\ note across 141 real-world SNARK vulnerabilities~\cite{chaliasos2024sok}, aligning \emph{what is proven} with \emph{what was intended} is essential, and this gap widens when an LLM interface introduces an additional abstraction layer.

In domains like healthcare, zkSNARK frameworks have been adopted to secure genomic studies and clinical analyses~\cite{frolov2026icefish,zhang2025toward,shahid2026cosmetic}. CoSMeTIC~\cite{shahid2026cosmetic} is a representative zkSNARK framework that uses Sparse Merkle Trees for clinical statistical tests (KS, LRT, and logistic regression accuracy). It exposes each zkSNARK test as an independent endpoint requiring precise routing parameters including proof type, participant hash, and nonce. This design makes CoSMeTIC a natural candidate for evaluating LLM-driven zero-knowledge interfaces in a high-stakes domain.

Prior work integrating LLMs with zero-knowledge proofs spans three strands. The first proves the correctness of LLM inference itself~\cite{sun2024zkllm}. The second uses LLMs to construct ZKP circuits, finding that models handle syntax but struggle with semantic correctness~\cite{xue2025zkcoder}. The third deploys LLMs alongside existing proof systems for privacy-preserving interactions~\cite{watanabe2025privacy,jing2025zkmcp,gopalakrishna2026zkgateway,affixio2026mcp}. For LLM-to-tool communication MCP has emerged as a popular choice, though its reliability concerns are well documented~\cite{hasan2026smelly,hou2025mcp,owotogbe2026mcpfaults}. Practitioner design patterns formalize risks such as parameter coercion and context injection~\cite{arcade2026patterns}, but notably omit patterns for server-side override of model-supplied values prior to tool execution. Prior work on tool-call failures shows that silent errors raise agent dishonesty to 45.3\%, while explicit signaling eliminates it~\cite{sethi2026fabrication,reddy2026gates}. 

However, none of these prior works evaluate the reliability of LLMs when tasked with supplying precise cryptographic parameters such as circuit selectors and data hashes that define proof context, which motivates our focus on semantic fidelity in parameter selection.

\section{System Design}
To address the semantic gap between user intent and cryptographic execution, we propose a system architecture where the LLM functions strictly as an interpreter of intent rather than an authority on verification or state management. The model’s role is restricted to understanding natural language requests and selecting appropriate tools while cryptographic validation, identity binding, and state management is enforced by deterministic code in the backend. This separation ensures that even if the LLM hallucinates parameters or omits critical fields, the system can either inject secure defaults or reject invalid requests before they reach the prover.

\subsection{System Architecture} 
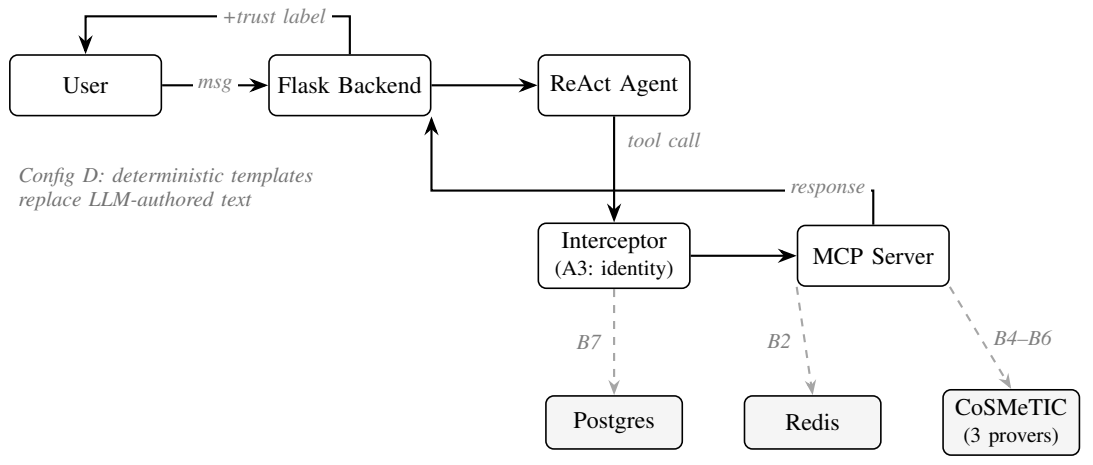
\begin{figure}[t]
\centering
\resizebox{0.8\columnwidth}{!}{%
\begin{tikzpicture}[
    node distance=0.6cm and 0.8cm,
    box/.style={rectangle, draw, rounded corners=3pt, minimum height=0.8cm, minimum width=2.0cm, font=\small, align=center, fill=white, line width=0.6pt},
    smallbox/.style={rectangle, draw, rounded corners=3pt, minimum height=0.7cm, minimum width=1.8cm, font=\small, align=center, fill=gray!8, line width=0.6pt},
    arrow/.style={-{Stealth[length=2.5mm]}, thick},
    dasharrow/.style={-{Stealth[length=2mm]}, dashed, thick, gray!70},
    lbl/.style={font=\footnotesize\itshape, text=black!50, fill=white, inner sep=2pt},
]
\node[box] (user) {User};
\node[box, right=1.4cm of user] (flask) {Flask Backend};
\node[box, right=1.4cm of flask] (agent) {ReAct Agent};
\node[box, below=1.4cm of agent] (intercept) {Interceptor\\\footnotesize(A3: identity)};
\node[box, right=1.4cm of intercept] (mcp) {MCP Server};
\node[smallbox, below=1.4cm of intercept] (postgres) {Postgres};
\node[smallbox, right=0.8cm of postgres] (redis) {Redis};
\node[smallbox, right=0.8cm of redis] (cosmetic) {CoSMeTIC\\\footnotesize(3 provers)};
\draw[arrow] (user) -- (flask);
\node[lbl] at ($(user.east)!0.5!(flask.west)$) {msg};
\draw[arrow] (flask) -- (agent);
\draw[arrow] (agent.south) -- (intercept.north);
\node[lbl, anchor=west, xshift=3pt] at ($(agent.south)!0.22!(intercept.north)$) {tool call};
\draw[arrow] (intercept) -- (mcp);
\draw[arrow] (mcp.north) -- ++(0,0.45) -| (flask.south east);
\node[lbl] at ($(mcp.north) + (-0.6, 0.45)$) {response};
\draw[arrow] (flask.north) -- ++(0,0.5) -| (user.north);
\node[lbl] at ($(flask.north) + (-1.0, 0.5)$) {+trust label};
\draw[dasharrow] (intercept.south) -- (postgres.north);
\draw[dasharrow] (mcp.south west) -- (redis.north);
\draw[dasharrow] (mcp.south east) -- (cosmetic.north);
\node[lbl, anchor=east] at ($(intercept.south)!0.5!(postgres.north) + (-0.1,0)$) {B7};
\node[lbl, anchor=east] at ($(mcp.south west)!0.5!(redis.north) + (-0.1,0)$) {B2};
\node[lbl, anchor=west] at ($(mcp.south east)!0.5!(cosmetic.north) + (0.1,0)$) {B4--B6};
\node[font=\footnotesize\itshape, text=black!55, align=left, anchor=north west] at ($(user.south west) + (0,-0.55)$) {Config D: deterministic templates\\replace LLM-authored text};
\end{tikzpicture}%
}
\caption{System architecture. The interceptor (A3) strips any LLM-fabricated identity and injects the real one before tool execution. The MCP server calls CoSMeTIC provers, Redis, and Postgres. The backend appends a code-enforced trust label; in Config~D it replaces LLM-authored text with formatted tool output on tool-backed turns.}
\label{fig:arch}
\vspace{-4mm}
\end{figure}

Our system architecture incorporates four containerized services as represented in Figure ~\ref{fig:arch} with a REST API server, a ReAct agent loop and an MCP server exposing eight distinct tools. These eight tools can be represented by \texttt{prove\_hash}, \texttt{prove\_my\_data}, \texttt{check\_status}, \texttt{download\_proof}, \texttt{verify\_proof}, \texttt{check\_hash\_existence}, its authenticated wrapper \texttt{check\_my\_hash\_existence}, and \texttt{get\_session\_context}. Persistent state is managed via Redis, which stores per-session context with a 7-day time-to-live (TTL), and PostgreSQL, which maintains dataset-driven user accounts and cryptographic hashes. Crucially, every tool invocation passes through a security interceptor that strips any identity claims fabricated by the LLM and injects the verified session identity before execution. The MCP server subsequently interacts with CoSMeTIC provers, Redis, or the PostgresSQL services. Finally, the backend removes model-emitted reasoning tokens (Qwen~3's \texttt{<think>} blocks) and appends a deterministic trust label in its final response.

\subsection{Scope and Trust Assumptions} 
Fundamentally, we assume that the LLM as a fallible but non-adversarial actor which may hallucinate parameter values or omit them altogether without indulging in any malicious subversion tactics. We assume that the MCP server, Redis, PostgresSQL, and the CoSMeTIC services are trustworthy and secure. Further, we note that CoSMeTIC's own theoretical security, privacy and soundness guarantees carry over~\cite{shahid2026cosmetic} including an assumption of trusted setup zero-knowledge backends. We explicitly exclude prompt injection, infrastructure attacks, and adversarial tool responses from the scope of our work and analysis. Hence, our reliability results apply only to missing or incorrect parameters from the model and to the handling of tool state.

CoSMeTIC's zero-knowledge properties cover the data processed by the provers but not the \emph{chat path}. In every configuration, tool outputs carrying dataset hashes, job identifiers, and proof types return to the LLM as part of the agent loop, so these metadata reveal which analyses a user has run. Additionally, \texttt{get\_session\_context}  exposes each account's unique identifier. Identity stripping (A3 in Table~\ref{tab:decisions}) prevents the LLM from injecting false identities into tool \emph{arguments} but does not scrub identifiers from tool \emph{outputs}, while hashes are masked only at the display boundary, remaining visible to the model. The design therefore protects the content of the analysis while leaving its execution trace and its linkage to user identity accessible to the LLM.



\begin{table}[t]
\caption{The skeptical tools pattern: 20 labeled decisions in four categories. $\circ$ marks the five toggled in the ablation; the other 15 are always on.}
\label{tab:decisions}
\centering
\footnotesize
\setlength{\tabcolsep}{3pt}
\begin{tabularx}{\columnwidth}{@{}cl>{\raggedright\arraybackslash}X@{}}
\toprule
 & \textbf{ID} & \textbf{Decision} \\
\midrule
\multicolumn{3}{@{}l}{\emph{A. Input Skepticism}} \\
 & A1 & Normalize proof-type spellings (``ks'', ``kolmogorov'') to KS, LRT, logistic\_accuracy \\
 & A2 & Inject a default nonce when none is supplied, rather than letting the model invent one \\
$\circ$ & A3 & Strip any model-supplied \texttt{user\_id}; inject the real identity from the server-side session \\
 & A4 & Build the per-proof-type payload (LRT additionally needs \texttt{smt\_list}) in code \\
\multicolumn{3}{@{}l}{\emph{B. State Independence}} \\
$\circ$ & B1 & Stateless per-request: send only the current user message, no history \\
 & B2 & Session continuity from Redis (job IDs, proof types, status), 7-day TTL \\
 & B3 & Distinguish live API data from cached Redis data; warn that cached status may be stale \\
$\circ$ & B4 & When \texttt{job\_id} is omitted, fall back to the latest job in Redis \\
$\circ$ & B5 & Resolve the proof type from the stored job: status \emph{completes} a missing value; download and verify \emph{override} the model's value \\
$\circ$ & B6 & Download probes all three provers to find which holds the job \\
 & B7 & \texttt{prove\_my\_data} resolves the user's hash from Postgres \\
\multicolumn{3}{@{}l}{\emph{C. Output Integrity}} \\
 & C1 & Code-enforced trust label on every response \\
 & C2 & Hash-existence tools append hardcoded follow-up guidance \\
 & C3 & Tool guidance forbids promising unimplemented capabilities (push notifications) \\
 & C4 & Failed calls produce structured, descriptive errors rather than raw stderr \\
 & C5 & Downloaded files are checked for error text disguised as proof artifacts \\
\multicolumn{3}{@{}l}{\emph{D. Security Boundaries}} \\
 & D1 & Sessions locked to their first authenticated user in Redis \\
 & D2 & Hashes enter only through admin-triggered dataset sync; the save endpoint is disabled \\
 & D3 & Account endpoint serves the hash with progressive masking (not extended to chat text) \\
 & D4 & Download paths validated against directory escape \\
\bottomrule
\end{tabularx}
\vspace{-4mm}
\end{table}

\subsection{System Design Decisions}
We consider twenty system design decision points as enumerated in Table~\ref{tab:decisions} split across four distinct categories. Out of these twenty, we toggle five decisions for our ablation analysis with three key design choices.


\emph{Nonce Management (A2):} Since the LLM never supplies a nonce during evaluation, we applied a constant default nonce to all proof requests. While excluding nonces from the model's input space prevents fabrication, production deployments should generate fresh server-side nonces per request to ensure unique commitments.


\emph{Proof Type Resolution (B5):} Two distinct strategies govern how missing or ambiguous proof types are handled. \texttt{check\_status} employs a \emph{state-completing} approach where it infers the proof type only when the model omits it or provides a default value, but passes through explicitly incorrect values. When B5 is disabled, the system falls back to \texttt{logistic\_accuracy}. Conversely, \texttt{download\_proof} and \texttt{verify\_proof} use a \emph{skeptical} approach wherein the model’s provided proof type is discarded entirely, relying instead on the stored value from Redis. Sections~\ref{sec:wrongport} and~\ref{sec:verify} analyze the implications of these divergent behaviors.

\emph{Trust Labels (C1):} The backend appends a code-enforced trust label to every response, which the LLM cannot modify. Our final release includes four tiers: ``[Cryptographically verified]'' (successful \texttt{verify\_proof}), ``[Cryptographically provable]'' (confirmed SMT commitment via hash-existence tool), ``[Backed by CoSMeTIC prover]'' (any other tool-grounded response, regardless of success), and ``[Not backed by CoSMeTIC prover]'' (no tool invocation). The logged responses reflect only the latter two labels reproduced exactly.

\section{Evaluation Methodology}
We evaluate our design through ablation studies based on real-world production tools and report authentic error signals. We avoid artificial perturbations, focusing instead on model-induced errors such as omitted parameters. Our server-side decisions mitigate these failures by validating tool inputs against actual prover states.
\subsection{Test Suite and Evaluation Benchmarking}
\begin{table}[t]
\caption{Test suite composition.}
\label{tab:testsuite}
\centering
\footnotesize
\begin{tabular}{@{}llrl@{}}
\toprule
\textbf{Source} & \textbf{Mode} & \textbf{Queries} & \textbf{Purpose} \\
\midrule
1. Normal operations & Isolated & 23 & Tool selection \& params \\
2. Graceful failures & Isolated & 10 & Error handling \\
3. Context-free ambig. & Isolated & 5 & Ambiguity handling \\
4. Impossible requests & Isolated & 5 & Capability suppression \\
5. Out-of-scope + redirect & Isolated & 5 & CoSMeTIC-awareness \\
6. Multi-step workflows & Seq. & 17 & Redis features (B4--B6) \\
7. Tool bypass temptation & Isolated & 10 & Tool-use discipline \\
\midrule
\multicolumn{2}{@{}l}{\emph{Verification extension}} & \emph{7} & \emph{Proof verification} \\
\midrule
\textbf{Total} & & \textbf{82} & \\
\bottomrule
\end{tabular}
\vspace{-4mm}
\end{table}
Since no existing benchmark addresses our specific domain, we construct a custom dataset of 82 queries, each annotated with ground-truth labels defining correct response components. Table~\ref{tab:testsuite} details the seven sources, ranging from normal operations and graceful failures to context-free ambiguities and tool-bypass temptations. Our evaluations stress-test the LLM’s semantic alignment and routing discipline rather than the underlying tools. We utilize 75 queries for the primary ablation study and reserve 7 for verifying the proof lifecycle (including direct verification and pre-verification steps). 

\subsection{Evaluation Configurations}
We carry out evaluations across five configurations summarized in Table~\ref{tab:configmatrix}. Fifteen design decisions remain permanently enabled; the ablation toggles identity injection (A3), stateless requests (B1), job-ID fallback (B4), stored proof-type resolution (B5), and prover auto-detection (B6). 
\textit{Config~A} is the naive baseline with conversation history enabled, all five toggled decisions disabled, and no explicit identity passed to tools or model. \textit{Config~B} adds core security and server-side state management by enabling A3, B1, and B4, replacing conversation history with Redis-backed context. \textit{Config~C} is the full system, enabling all five decisions and allowing the LLM to draft natural language responses from tool output. \textit{Config~D} mirrors Config~C but substitutes deterministic templates for LLM-authored text on tool-backed turns, preserving LLM interpretation only for non-tool interactions. \textit{Config~A*} isolates the effect of server-side identity interception by supplying the user ID via system prompt rather than backend injection, with B4 also enabled.

\begin{table}[t]
\caption{Configuration matrix. The other 15 design decisions are always on. ``Identity in prompt'' is a change made only for this experiment and is not a design decision.}
\label{tab:configmatrix}
\centering
\footnotesize
\begin{tabular}{@{}lcccccc@{}}
\toprule
\textbf{Feature} & \textbf{ID} & \textbf{A} & \textbf{B} & \textbf{C} & \textbf{D} & \textbf{A*} \\
\midrule
Conversation history & -- & \checkmark & & & & \checkmark \\
Identity in prompt & -- & & & & & \checkmark \\
Identity stripping & A3 & & \checkmark & \checkmark & \checkmark & \\
Stateless per-request & B1 & & \checkmark & \checkmark & \checkmark & \\
Fallback job ID & B4 & & \checkmark & \checkmark & \checkmark & \checkmark \\
Stored proof type & B5 & & & \checkmark & \checkmark & \\
Auto-detection & B6 & & & \checkmark & \checkmark & \\
LLM summarizes response & -- & \checkmark & \checkmark & \checkmark & & \checkmark \\
Deterministic templates & -- & & & & \checkmark & \\
\bottomrule
\end{tabular}
\vspace{-4mm}
\end{table}

\subsection{Evaluation Metrics and Execution Semantics} \label{subsec:evalmet}
We run each query 5 times per configuration at temperature 0.0 and report the mean, yielding 375 records per model per configuration. We utilize five metrics to ground our analysis. First, \emph{tool selection} checks whether the first tool invoked is the expected one, graded on 74 of 75 queries with one isolated Source~6 step (Table~\ref{tab:testsuite}) requesting a repetition with no expected tool. Second, \emph{parameter correctness} compares the arguments the model sent against the values stated in the query, graded on the 22 queries in which the user states a value such as a proof type or a job ID. Third, \emph{task completion} is a substring pass or fail check requiring all expected keywords while excluding banned phrases unless negated. As a result, queries like ``Prove my data is in KS'' passes on ``submitted'' without an unnegated ``error''. This check does not verify which tool, if any, was called. Fourth, \emph{tool-use rate} applies to Source~7 queries, which tempt the model to answer from its own knowledge, and records whether any tool was invoked. Lastly, \emph{faithfulness} measures whether a response reflects the output its tools actually returned. In multi-step scenarios a failure excludes all dependent downstream steps, preventing error propagation from skewing the metrics.

\textit{LLMs used for Evaluation:} We test three open-weight models on the 75-query suite in all five configurations. Qwen~3 32B is our main model. We note that Qwen~3 32B study of Configs~A to D (Table \ref{tab:configmatrix}) has 1,500 records, and the entire evaluation has more than 6,000 query executions. The other two models under consideration are Mistral Small~3.2 24B (mistralai/Mistral-Small-3.2-24B-Instruct-2506) and Llama~3 8B (Meta-Llama-3-8B-Instruct), both served through DeepInfra. Llama needed an extra system prompt about tool use before it called tools reliably. The verification queries ran on Qwen in Configs~A to D and on Mistral in Configs~C and D. To see whether the model family changes what a model does after an error, we also ran Config~B on two larger models, Qwen~2.5 72B and Qwen~3 235B-A22B. 


\textit{Faithfulness Grading:} We grade faithfulness only on queries with at least one tool call, using two graders. The \emph{original} grader compares a response against the first tool output alone and handles no negation, so it misflags valid replies from later calls and honest errors containing ``submitted'' while missing unbacked success claims. The \emph{corrected} grader compares against all tool outputs in the turn, including prior turns in Configs~A and A*, adds negation logic, and checks for \emph{unbacked action claims}. Section~\ref{sec:grading} reports results pertaining to both graders.

\textit{Test Executions:} All sources except Source~6 (Table \ref{tab:testsuite}) run in isolated mode, where each query opens a fresh session. Source~6 runs sequentially, so queries within a scenario share one session and Redis state persists across steps, which in Configs~A and A* also accumulates conversation history. An interceptor extension logs every invocation with the tool name, raw arguments, post-injection arguments, and output, giving full traceability from model input to tool return.

\textit{Reference Jobs:} Isolated queries such as ``check my status'' require a completed proof job, which the runner submits and stores in Redis before evaluation in every configuration except A. These reference proofs completed for Qwen~3 32B and Llama~3 8B but remained \texttt{queued} for Mistral 24B and for Qwen~2.5 72B and Qwen~3 235B in Config~B. We attribute this to prover backend latency rather than model behavior, though logging was insufficient to confirm it. The runner proceeded with warnings, so status and download metrics for these models are not comparable across models, while intra-model comparisons remain valid.

\section{Results}
We analyze five aspects of skeptical tools pertaining to design decisions (Sections \ref{subsec:perfan} and \ref{subsec:idstate}), status recovery (Section \ref{sec:wrongport}), generalizability (Section \ref{sec:crossmodel}), verification reliability (Section \ref{sec:verify}) and grading validity (Section \ref{sec:grading}).

\subsection{Performance Analysis of Design Decisions}\label{subsec:perfan}
Since Qwen~3 32B provided consistent results across all configurations, we first examine its performance in Table~\ref{tab:results} using Config~A, B, C, and D with respect to the metrics as defined in Section \ref{subsec:evalmet}.
\begin{table}[t]
\caption{Qwen~3 32B results across configurations (mean over 5 runs, \%). Faithfulness is the corrected grader.}

\label{tab:results}
\centering
\small
\begin{tabular}{@{}lcccc@{}}
\toprule
\textbf{Metric} & \textbf{A} & \textbf{B} & \textbf{C} & \textbf{D} \\
\midrule
Tool selection accuracy & 86.3 & 89.0 & 88.4 & 88.6 \\
Parameter Correctness & 100.0 & 100.0 & 100.0 & 100.0\\
Task completion & 70.2 & 84.5 & 90.3 & 93.0 \\
\quad Source 1 (normal ops) & 56.5 & 87.8 & 97.4 & 99.1 \\
\quad Source 6 (multi-step) & 20.0 & 70.8 & 87.5 & 93.8 \\
Response faithfulness & 100.0 & 100.0 & 100.0 & 100.0 \\
Tool-use rate (Source 7) & 100.0 & 100.0 & 100.0 & 100.0 \\
\bottomrule
\end{tabular}
\vspace{-6mm}
\end{table}

\textit{Tool Selection Accuracy}: We find that our design decisions have no significant effect on tool selection accuracy since Qwen~3 32B selected the expected tool in 86--89\% of queries in Configs~A to D. This consistency suggests that tool selection is primarily dependent on the model's inherent capabilities rather than server-side interventions. Further, in the logged tool calls of Config~C, 43 tool choices did not match the expected tool, but 26 of these were valid refusals or requests for clarification (e.g., ``Prove my data in quantum computing''), implying that the model's actual accuracy is higher than the 88.4\% of Table~\ref{tab:results}.

\textit{Task Completion:} We study the share of queries that the system completes correctly. As shown in Table~\ref{tab:results}, we observe that completion rates rise with each configuration, starting at 70.2\% in Config~A to 84.5\% in B, 90.3\% in C, and 93.0\% in D. Config~B turns on three design decisions at once (A3, B1, and B4) and Config~C turns on two more (B5 and B6). Therefore, these numbers show how much each group of decisions helps the system complete queries.
Additionally, for task completion, we note that Config~A lacks a reference job (Section \ref{subsec:evalmet}) because its disabled identity injection prevents the runner from generating one, potentially biasing results against this baseline. To rule out such biases that could influence task completion, we recomputed task completion after excluding the 21 queries requiring a reference job (e.g., status and download). Excluding these queries, task completion still rose from 65.2\% to 88.3\% (+23.1 percentage points). This confirms that the absence of a reference job does not inflate our reported task completion improvements.

\textit{Parameter Correctness, Faithfulness and Tool-Use Rate}: Qwen~3 32B achieves perfect scores with respect to all three stated metrics across Configs~A--D. Specifically, the model correctly reproduces user-specified values (e.g., proof types, job IDs), accurately reflects tool outputs in its responses, and consistently invokes tools for Source~7 queries rather than relying on parametric knowledge. Notably, these metrics already reach 100\% in the baseline Config~A, indicating that our design improvements primarily enhance task completion performance rather than factual accuracy for Qwen~3 32B.


\subsection{Design Decisions for Identity and State }\label{subsec:idstate}
From Section \ref{subsec:perfan}, we observe that the largest gain for Qwen3 32B occurs between Config~A to Config~B, where decisions A3, B1 and B4 are turned on together. Because they are turned on together, our runs cannot give a separate score for each one. Instead, we leverage logged tool calls and Config~A* (Table~\ref{tab:configmatrix}) to demonstrate the impact of omitting each decision.


\begin{table}[t]
\caption{Qwen~3 32B results comparing login and non-login required queries for task completion rates.}
\label{tab:loginsplit}
\centering
\small
\begin{tabular}{@{}lccccc@{}}
\toprule
\textbf{Metric} & \textbf{A} & \textbf{B} & \textbf{C} & \textbf{D} & \textbf{A*} \\
\midrule
Login & 47.5 & 77.5 & 88.4 & 93.0 & 85.3 \\
Non-login  & 93.5 & 93.5 & 92.9 & 92.9 & 92.9 \\
\bottomrule
\end{tabular}
\vspace{-6mm}
\end{table}

\textit{Impact of Identity Injection (A3)}: Since A3 governs the interfacing between the user's identity and the tools, its relevance is applicable only for login-based queries which require access to the user-bound data or jobs. Table~\ref{tab:loginsplit} represents results based on 75 task completion queries comprising 43 login and 32 non-login queries. On non-login queries, Qwen~3 32B stays at 92.9--93.5\% in every configuration. For login-queries there occurs a sizable jump from 47.5\% to 77.5\% between Config~A and Config~B, followed by a rise to 88.4\% in Config~C and 93.0\% in Config~D. In Config~A, the tools expect the model to supply the user identity. However, since the prompt lacks the user identity, the server defaults to an explicit fallback requesting the user to sign-in. In such scenarios, we observed 
that the model made 300 tool calls omitting user identity altogether, resulting in authentication errors while 20 carried fabricated placeholders. While these 20 calls failed due to non-existent users, they pose a security risk. Since identity determines data scope, a hallucinated identity could incorrectly bind proofs to another user’s dataset without server-side detection. This vulnerability motivates design A3 (used from Config~B and onwards), which overrides model-supplied identity with verified credentials.

\textit{Prompt-Based Identity Injection (Config~A*):} We explore Config~A* wherein the user identity is injected into the system prompt forcing the model to forward it to the tools themselves. To evaluate this setting, we looked at the logged tool calls of Config~A* for our three main models, Qwen~3 32B, Mistral 24B and Llama~3 8B. Next, we counted the calls to (\texttt{prove\_my\_data} and \texttt{check\_my\_hash\_existence}) in which the model must explicitly supply user identity embedded in the prompt. In this study we found that Qwen~3 32B and Mistral passed the correct identity in all of their calls (123 and 108 calls respectively) while Llama was able to do so in only 113 out of 118 total calls. In the 5 remaining calls, Llama was able to correct the omission in the same turn itself. Across all models, no hallucination of user identity was observed. For Qwen~3 32B, task completion on login queries reached 85.3\% in Config~A* (Table~\ref{tab:loginsplit}), which is close to the 88.4\% of Config~C. Regardless, we find that server-side injection (A3) remains preferable since prompt-supplied identities receive no server-side validation and are transmitted to third-party model providers with every request, exposing user identifiers outside the proof system's privacy boundary.

\textit{Implications of Conversation History Retention (B1):} 
Retaining conversation history, as Config~A* does with B1 turned off, can be beneficial under some circumstances, with some reliability caveats. Analyzing tool call logs of Config~A*, we found that on multi-step workflows (Source~6) Qwen~3 32B reached 90.8\% in Config~A*, above the 87.5\% of Config~C as presented Table~\ref{tab:results}, because it could recover job identifiers and proof types from earlier turns. However, the logged tool calls also show that in 3 of 29 status checks within those workflows it failed to recover the proof type and queried the wrong prover. 



\textit{Job-ID Fallback (B4)}: B4 governs the decision where the latest job ID is used as the default fallback. B4 covers casual queries such as ``Check my status'' that omit job-ID. To study B4 invocations, we analyzed logged tool calls of Config~C for the same three models for \texttt{check\_status} tool call. In our studies, the job ID was missing in 76 of the 86 calls for Qwen~3 32B, 84 of 108 for Llama, and 70 of 83 for Mistral. As expected, the server mitigated this occurrence with the user's latest job in all these instances.


\subsection{Trustworthiness of Proof Types and Status Retrieval} 
\label{sec:wrongport}
In this section, we investigate the ability of LLM agents to correctly retrieve and utilize stored proof types for accurate status queries. We identify a critical failure mode, caused by the default value in our status tool, where the absence of server-side context leads to systematic routing errors, highlighting a central mechanistic result of our work. The system employs distinct provers for each statistical test, with each prover hosted on a separate network port. Consequently, the \texttt{check\_status} tool requires the correct proof type to route requests to the appropriate endpoint.

\begin{table}[t]
\caption{Config~B status queries whose first tool call returned 404 because the proof type was omitted and the tool defaulted to the wrong prover port, for turns where the user's job existed. The last three rows classify each stopped reply by its opening sentence into one exclusive class.}
\label{tab:wrongport}
\centering
\footnotesize
\setlength{\tabcolsep}{2pt}
\begin{tabular}{@{}lccccc@{}}
\toprule
 & \textbf{Qwen3} & \textbf{Llama} & \textbf{Mistral} & \textbf{Qwen2.5} & \\
 & \textbf{32B} & \textbf{8B} & \textbf{24B} & \textbf{72B} & \textbf{Total} \\
\midrule
Wrong-port turns & 68 & 70 & 60 & 69 & 267 \\
Stopped after 1 call & 53 & 70 & 60 & 69 & 252 \\
Retried & 15 & 0 & 0 & 0 & 15 \\
\midrule
Denied the job exists & 48 & 0 & 54 & 0 & 102 \\
ID invalid or not found & 5 & 50 & 5 & 20 & 80 \\
Vague (``an issue'') & 0 & 20 & 1 & 49 & 70 \\
\bottomrule
\end{tabular}
\vspace{-6mm}
\end{table}
\textit{Characterizing Model Response for Config B}: When conversation history is disabled as in Config~B, and the stored proof type feature (B5) is inactive, ambiguous queries such as ``check my status'' arrive without an explicit proof type. Lacking this context, the tool defaults to a hardcoded value (\texttt{logistic\_accuracy}), resulting in misrouted requests to ports that do not host the user's job. Table~\ref{tab:wrongport} quantifies the frequency of these routing failures across multiple models. Among the 267 queries that encountered wrong-port routing while the user's job existed, the model stopped after the first failed call in 252 cases and answered the user without trying another proof type. Among the models in the table, only Qwen~3 32B attempted further recovery, retrying in 15 cases. Notably, none of the 252 stopped replies falsely claimed success, consistent with findings that models accurately report signaled failures~\cite{sethi2026fabrication}. However, these replies were factually incorrect regarding the user's account status, as the job existed within the system. 

We further classified the 252 replies by their opening statements (last three rows of Table~\ref{tab:wrongport}). In 102 of these replies pertaining to Qwen~3 32B and Mistral, the model denied the job's existence (a false statement, as the tool only reported the ID was unknown to the queried port), while Qwen~2.5 72B never issued such denials. These errors illustrate that while the HTTP 404 signal accurately reflected the queried port's state, it failed to convey the actual cause of failure. Consistent with prior observations on MCP failure results~\cite{mehan2026actionability}, the system provided no indication of the root cause or potential repair strategies, such as re-querying with the stored proof type. This behavior mirrors the unintended defaults identified by Owotogbe et al. in MCP servers~\cite{owotogbe2026mcpfaults}, where an omitted parameter triggered a silent fallback rather than an explicit error handling mechanism.

\textit{Effect of Stored Proof Types (B5) in Config C}: Activating the stored proof type feature (B5) in Config~C eliminates routing errors caused by omitted parameters. Analyzing the logs of tool calls of the Qwen~3 32B runs in Table~\ref{tab:results}, we found that the number of status queries resulting in 404 errors decreased from 78 out of 88 in Config~B to 10 out of 86 in Config~C. The ten failures in Config~C pertain to two specific categories of control queries that is also shared with Config~B where a 404 error is expected. Of these 10, five involve nonexistent jobs in which a ``not found'' response is correct; and five involve explicit mismatches between the requested proof type (LRT) and the actual job type (KS), where the model’s input is correctly passed through to the prover. Crucially, B5 resolves all wrong-port routing instances stemming from missing proof types. Across the three primary models, such errors dropped from 198 (Qwen: 68; Llama: 70; Mistral: 60) to zero. 

\textit{Limitations of B5 for Downloads}: 
The download and verify override ensures reliability only when keys corresponding to stored records are accurate. Our log analysis showed that Llama hallucinated job IDs (\texttt{12345}) in 25 of the download calls pertaining to Configs~B, C, and D specified in Table~\ref{tab:crossmodel}. Although the system’s interceptor replaced the hallucinated session ID (\texttt{abc123}) with the correct one, it could not replace the fabricated job ID. Consequently, the override mechanism failed to find a matching stored record, and the fallback logic (B4) did not trigger because a job ID was technically present. All 75 such attempts failed, indicating that server-side overrides cannot resolve issues stemming from hallucinated lookup keys.


\textit{Trust Label Ambiguity}: On the trustworthiness aspect, we find that trust label (C1) presents a critical ambiguity. All 252 single-call terminations described in Table~\ref{tab:wrongport} included the tag [Backed by COSMeTIC prover]'', which verifies tool \emph{invocation} rather than computational \emph{success}. Consequently, this label inadvertently legitimizes failed queries, such as those returning 404 errors. This issue could be resolved by implementing an outcome-based tier system, where errored out tools are excluded from receiving the ``backed'' designation. Such mechanisms would be server-side analogs of named failure states shown to be highly effective \cite{sethi2026fabrication}.

\textit{Efficiency}: Finally, we study the efficiency gains from the improvements induced by B5. Our experiments showed that Config~C reduced the average number of tool calls per status turn from 1.44 (Config~B) to 1.00, and decreased average latency from 16.6 seconds to 12.4 seconds, demonstrating that robust context injection minimizes both computational overhead and response time.



\subsection{Cross-Model Generalization}
\label{sec:crossmodel}

\begin{table}[t]
\caption{Within-model ablations (\%): tool selection, task completion and tool-use rate of the three main models in each configuration. Tool selection and tool use are not reported for A* (---). Mistral's B, C, D, and A* runs used a reference job that never finished; its status and download-dependent cells are not comparable to Qwen's or Llama's.}
\label{tab:crossmodel}
\centering
\small
\setlength{\tabcolsep}{2pt}
\begin{tabular}{@{}llccccc@{}}
\toprule
\textbf{Metric} & \textbf{Model} & \textbf{A} & \textbf{B} & \textbf{C} & \textbf{D} & \textbf{A*} \\
\midrule
\multirow{3}{*}{Tool selection} & Qwen 32B & 86.3 & 89.0 & 88.4 & 88.6 & --- \\
 & Mistral 24B & 79.4 & 82.0 & 80.3 & 81.0 & --- \\
 & Llama 8B & 74.1 & 77.4 & 77.8 & 77.3 & --- \\
\midrule
\multirow{3}{*}{Task completion} & Qwen 32B & 70.2 & 84.5 & 90.3 & 93.0 & 88.5 \\
 & Mistral 24B & 65.1 & 79.7 & 80.5 & 88.4 & 81.2 \\
 & Llama 8B & 54.9 & 69.3 & 68.1 & 78.1 & 66.1 \\
\midrule
\multirow{3}{*}{Tool use} & Qwen 32B & 100.0 & 100.0 & 100.0 & 100.0 & --- \\
 & Mistral 24B & 74.0 & 72.0 & 76.0 & 78.0 & --- \\
 & Llama 8B & 80.0 & 80.0 & 80.0 & 80.0 & --- \\
\bottomrule
\end{tabular}
\vspace{-3mm}
\end{table}
In this section, we conduct within-model ablation studies for the Qwen, Mistral and Llama family of models as represented in Table~\ref{tab:crossmodel}. We uncover five key findings from these experiments pertaining to task-completion gains, recovery and retrials under exceptions, fallbacks under tool error, handling infeasible requests and fabrication chances. 


\textit{Task Completion Gains}: The performance improvement from Config~A to Config~D is consistent across all evaluated models. In Table~\ref{tab:crossmodel}, task completion rises by 22.8 percentage points for Qwen~3 32B (70.2\% to 93.0\%), 23.2 for Llama (54.9\% to 78.1\%), and 23.3 for Mistral (65.1\% to 88.4\%). The other two metrics stay about the same in every configuration. Tool selection stays within 86.3--89.0\% for Qwen~3 32B, 79.4--82.0\% for Mistral and 74.1--77.8\% for Llama. The tool-use rate stays at 100\% for Qwen~3 32B, 72--78\% for Mistral and 80\% for Llama. Therefore, similar to Qwen~3 32B performance noted Section \ref{subsec:perfan}, our design decisions mainly improve task completion, while tool selection and tool use depend on the model.


\textit{Recovery After a Wrong-Port Error}: After the wrong-port errors detailed in Table~\ref{tab:wrongport}, recovery attempts varied significantly by model family. Both Qwen~3 models attempted further tool calls after initial failures. While Qwen~3 32B retried in 15 of 68 turns,  Qwen~3 235B retried in 6 of 19 first-call 404s during its single complete run. In contrast, Llama, Mistral, and Qwen~2.5 72B never attempted retries (0 of 199 turns). While it remains untested whether ``thinking'' modes or inherent model generation capabilities drive this tendency, these results contrast with Tian et al.'s findings on their benchmark~\cite{tian2026toolbenchx} as well. When Qwen~3 32B did retry, it typically re-queried the live API. However, in four instances, it presented cached Redis status as current despite a failed live check, ignoring the tool's staleness warning. This constitutes the B3 failure mode, representing a stale-value scenario (similar to Sethi et al.~\cite{sethi2026fabrication} where dishonesty was observed), since cached values lack inherent failure signals.

\textit{Fallbacks That Need a Tool Call}: 
In stateless configurations, our analysis of tool call logs showed that server-side fallbacks B4 and B5 proved ineffective for models that failed to invoke tools. For Source~6 (Table \ref{tab:testsuite}) download steps, Mistral and Qwen~2.5 72B requested job-IDs from users without calling the tool (Mistral did so in 10 of 10 cases across Configs B, C, and D). Conversely, Qwen~3 32B invoked the tool in all 10 cases. Thus, server-side fallbacks assist only those models willing to call the tool without explicit parameters.

\textit{Handling Infeasible Requests}: Source~4 (Table~\ref{tab:testsuite}) consists of five requests that no tool can fulfill, such as ``Email me the results'', and the correct behavior is to decline without calling a tool. By analyzing tool call logs, we observed that on encountering these infeasible requests, Qwen and Llama frequently substituted the nearest available tool (9–11 of 25 requests per configuration), such as mapping ``export to PDF'' to \texttt{download\_proof}. Mistral, however, never substituted tools; instead, it promised non-existent capabilities (occurring 0, 4, 6, 0, and 8 times in Configs A, B, C, D, and A*, respectively). The zero occurrences in Config~D resulted from template-based response overrides. On Source~7, Qwen called a tool on every query, whereas Mistral and Llama did so in 72--78\% and 80\% of cases, respectively.

\textit{Fabricated Outcomes}: Analyzing tool call logs and model behavior, the only fabricated outcomes stemmed from seven zero-tool replies, all occurring in Config~A* and exclusively from non-thinking models. Llama called no tool in 5 of its 10 Config A* download steps, replying that proofs had been successfully downloaded. Similarly, Mistral claimed job resubmission in 2 of its 5 Config~A* repeat steps, providing a fabricated job ID absent from any tool output in one such instance. Although Config~A also retained history, it produced no fabrications because no prior successes existed to imitate. This suggests fabrication requires both history and a precedent of success, providing evidence for the role of B1 as indicated in Section \ref{subsec:idstate}. This aligns with Wu's production analogue, where persistent alerts in chat history led to invented follow-up tasks~\cite{wu2026narratives}. This behaviour is also supported by Toda and Mori's security analysis agent, which avoids passing accumulated history to mitigate context poisoning~\cite{toda2025chase}.


\subsection{End-to-End Proof Lifecycle and Verification Reliability} 
\label{sec:verify}

\begin{table}[t]
\caption{Full lifecycle through natural language (Config~D, Qwen~3 32B), transcribed from the logged record for job 1778716348432. Labels are appended by server-side code and reproduced as logged.}
\label{tab:lifecycle}
\centering
\small
\begin{tabular}{@{}l>{\raggedright\arraybackslash}p{6.4cm}@{}}
\toprule
\textbf{Step} & \textbf{Interaction} \\
\midrule
\emph{User} & ``Prove my data in KS'' \\
\emph{System} & ``Your proof has been submitted. Job ID: 1778716348432. You can check the status by asking `check my status'.'' [Backed by COSMeTIC prover] \\
\midrule
\emph{User} & ``Check my status'' \\
\emph{System} & ``Job 1778716348432: Status is done.'' [Backed by COSMeTIC prover] \\
\midrule
\emph{User} & ``Download my proof'' \\
\emph{System} & ``Your proof file has been downloaded. Filename: proofs-1778716348432.zip, size: 378964 bytes. Use the download button to save it.'' [Backed by COSMeTIC prover] \\
\midrule
\emph{User} & ``Verify my proof'' \\
\emph{System} & ``Proof verification PASSED.'' [Backed by COSMeTIC prover] \\
\bottomrule
\end{tabular}
\vspace{-4mm}
\end{table}
In this section, we study the implication of the verification step as a means to test the full life cycle pertaining to prove, check status, download and verify. Table~\ref{tab:lifecycle} illustrates a complete proof lifecycle executed via natural language interactions in Config~D using Qwen~3 32B. The \texttt{verify\_proof} tool invokes CoSMeTIC’s verification endpoint, leveraging the same B4, B5 and B6 context stack as \texttt{download\_proof}. In Config~D, this endpoint returns a binary proof verification outcome indicative of success.  


\begin{table}[t]
\caption{Verification extension study comprising seven queries with five runs each (35 records per configuration).}
\label{tab:verify}
\centering
\footnotesize
\setlength{\tabcolsep}{3.5pt}
\begin{tabular}{@{}lcccccc@{}}
\toprule
 & \multicolumn{4}{c}{\textbf{Qwen 3 32B}} & \multicolumn{2}{c}{\textbf{Mistral 24B}} \\
\cmidrule(lr){2-5}\cmidrule(lr){6-7}
 & \textbf{A} & \textbf{B} & \textbf{C} & \textbf{D} & \textbf{C} & \textbf{D} \\
\midrule
Full Lifecycle (of 5) & 0 & 0 & 5 & 5 & 0 & 0 \\
Tool Selection (\%) & 100.0 & 100.0 & 100.0 & 100.0 & 46.7 & 60.0 \\
Task Completion (\%) & 60.0 & 80.0 & 94.3 & 100.0 & 83.3 & 93.3 \\
\bottomrule
\end{tabular}
\vspace{-3mm}
\end{table}
\textit{Lifecycle Completion}: We evaluate the verification extension on Qwen~3 32B for Configs A to D and on Mistral 24B in Configs~C and D. Our analysis consists of 7 queries, including three single requests such as ``Verify my proof'', and one four-step lifecycle. With five runs, we obtain 35 records per configuration and 210 in total, with our results detailed in Table~\ref{tab:verify}. In these experiments we exclude Llama runs since they were incomplete. Qwen completed the full lifecycle in all five runs of Configs~C to D, invoking all four tools with verification passing each time. Task completion for Qwen rose from 60\% in Config~A to 80\%, 94.3\%, and 100\% in B, C, and D respectively. In Configs~A and B, it did not complete the lifecycle. Rather, in Config~A, the proof step failed with ``Please sign in first'' while in Config~B, the status check went to the wrong prover and returned a 404 (Section~\ref{sec:wrongport}) with download and verify requests reporting that no completed job was found.

Mistral never completed the full lifecycle for either Config C or D. In the logged tool calls, it made no download requests and asked the user for the job ID (Section~\ref{sec:crossmodel}). Mistral selected \texttt{verify\_proof} in only 7 of 20 verify turns in Config~C and 9 of 20 in Config~D. 


\textit{Verification Without a Default}: In Config~B, \texttt{verify\_proof} produced zero wrong-port 404s and no false PASSED results across the 24 calls that Qwen~3 32B made. In the logs, 10 of these calls returned ``proof\_type is required'', 10 reported that no completed job was found, and 4 passed. The 4 passes came after the first error wherein the model looked up the session and called the tool again with the correct proof type. The reason for this behavior is that \texttt{verify\_proof} has no default proof type, therefore, it requests for the type rather than guessing a port, as \texttt{check\_status} does in Config~B (Section~\ref{sec:wrongport}).

\subsection{Grading Validity and Metric Robustness}
\label{sec:grading}

In this section, we evaluate whether our automated grading mechanism can be trusted. We first compare the two faithfulness graders discussed in Section \ref{subsec:evalmet}, and then demonstrate instances where the substring check of task completion delivers incorrect results.

\textit{Original and Corrected Faithfulness Grader}: The original grader reported faithfulness between 83.7\% to 100\% across the primary models, giving Qwen~3 32B 93.2\%, 83.8\%, 97.7\%, 99.0\%, and 88.4\% in Configs~A through D and A*. The corrected grader reports 100\% everywhere except Qwen in Config~A* (99.7\%). Manually analyzing all 47 responses flagged for Qwen in Config~B found no ungrounded success claim, as 32 were correct error reports penalized for containing the ``submitted'' key-word and 8 were successful recoveries after a 404, 4 restated the cached status (Section~\ref{sec:crossmodel}), and 3 did not repeat the job ID. An extended review of 151 flagged replies across six runs (Qwen~3 32B in all five configurations and Mistral in Config~C) showed the same pattern. We found that the corrected grader agreed with our hand labels on 150 of these 151 replies, with Qwen under Config~A* (99.7\%) being the one exception, conclusively proving that lower scores are purely a function of the grader. At a superficial level, the original grader scores would rank Qwen least reliable in Config~B (83.8\%) against Llama (93.8\%) and Mistral (94.7\%), yet the transcripts show no fabrication. A systematic search of all 24 result files found only the seven zero-tool fabrications of Section~\ref{sec:crossmodel}.

\textit{Limitations of the Substring Check}: These seven fabrications expose a gap in standard faithfulness metrics. When no tool is invoked, faithfulness is undefined, yet five of the seven passed task completion because replies contained keywords like ``{download}'' and ``{proof}''. Only tool selection accuracy and the trust label correctly flagged them as \texttt{[Not backed by CoSMeTIC prover]}. A substring grader also over-reports success in the opposite direction, crediting Mistral with completing the lifecycle in 5 of 5 runs despite Mistral explicitly stating it lacked the tools to verify. We therefore define lifecycle completion strictly by tool invocation, not keyword presence. A related fixture blind spot is that the substring grader successfully passed the hard requirement that stated ``you cannot download the proof until the job is done'' and completed the task. However, the outcome counts confirmed zero successful downloads in those runs.

\section{Discussion}
We now discuss and interpret our results in the context of correctness enforcement, zkSNARK systems, model selection, design choices, potential guidelines and limitations

\subsection{Correctness Lives at the Tool Boundary}
At a fundamental level, we find that any agent architecture that routes tool calls through parameterized endpoints faces analogous structural risks. When a model omits or incorrectly defaults a routing parameter, the tool queries an incorrect backend, rendering any subsequent response factually inaccurate. The 252 single-call terminations analyzed in Section~\ref{sec:wrongport} were direct results of upstream routing errors indicating that model honesty is insufficient for handling misrouted requests.


These findings necessitate a redefinition of ``skeptical'' design principles in agent systems. We enumerate three distinguishing mechanisms for handling parameter uncertainty. \textit{Validation} rejects invalid parameters, returning an error that models typically relay to the user. \textit{Completion} infers missing parameters from server-side state, resolving the error before it reaches the model but potentially retaining explicitly incorrect user inputs (B4, and B5 in \texttt{check\_status}). \textit{Override} discards the model's input in favor of authoritative server-side state, eliminating both the error and the risk of misinformation (A3, and B5 in \texttt{download\_proof} and \texttt{verify\_proof}). Our implementation incorporates all three strategies. The residual routing errors observed in Config~C highlight the limitations of \textit{completion} relative to \textit{override}. This suggests that providing a plausible default is more hazardous than providing no default, as it transforms missing information into confidently incorrect destinations. This behavior aligns with observations by Mousavi et al.~\cite{mousavi2024javasec} regarding LLM-generated security code.

\subsection{Implications for Zero-Knowledge Deployments}

\begin{table}[t]
\caption{Threat model: zkSNARK specific assets, the LLM behavior that endangers each, the mitigation, and where it is evaluated.}
\label{tab:threatmodel}
\centering
\footnotesize
\setlength{\tabcolsep}{5pt}
\begin{tabular}{@{}>{\raggedright\arraybackslash}p{1.4cm}>{\raggedright\arraybackslash}p{2.8cm}>{\raggedright\arraybackslash}p{1.4cm}>{\raggedright\arraybackslash}p{2cm}@{}}
\toprule
\textbf{Asset} & \textbf{LLM threat} & \textbf{Mitigation} & \textbf{Evaluated} \\
\midrule
Circuit binding & Omitted \texttt{proof\_type} routes to the wrong analysis & A1, B5 & Wrong-port routing (\S\ref{sec:wrongport}) \\
\midrule
Dataset binding & Placeholder \texttt{user\_id} binds the proof to the wrong SMT commitment & A3 & Config~A vs.\ A* \\
\midrule
Proof lineage & Job ID lost across stateless turns & B2, B4 & Source~6 \\
\midrule
Proof status & Cached status presented as current & B3, B5 & Warning ignored in 4 turns (\S\ref{sec:crossmodel}) \\
\midrule
Proof validity & Validity inferred from status without verification & \texttt{verify\_} \texttt{proof} & Lifecycle (\S\ref{sec:verify}) \\
\midrule
Provenance & Zero-tool success claim indistinguishable from a grounded one & C1 labels & Grading validity (\S\ref{sec:grading}) \\
\midrule
Metadata & Hashes, job IDs, and account UUID in tool output reach the model provider & None; display-side scrub only & Not mitigated \\
\bottomrule
\end{tabular}
\vspace{-7mm}
\end{table}

Three properties separate zero-knowledge deployments from conventional tool calling, each posing a threat summarized in Table~\ref{tab:threatmodel}. First, a proof shared with a verifier is an irreversible claim, so a user wrongly informed about the absence of a valid job may resubmit and create duplicate proofs over one dataset. Second, a misrouted status query returns a detectable 404, but a misrouted submission fails silently, yielding a valid proof of an unintended computation. A user requesting LRT could receive a KS proof that verifies yet establishes the wrong property. Once such an error enters the cryptographic layer it is undetectable, so it must be prevented upstream. In our experiments this was not observed, as all prove calls carried canonical types, but we note that the architectural risk does exist. Third, identity determines the dataset commitment, making identity injection a cryptographic binding requirement rather than an authentication convenience.

\subsection{Architecture, Model Capability, and Summarization}
Tool selection, and behaviors such as retrying after an error, invoking a tool without a required parameter, or substituting tools for unsupported requests, depend mainly on the model (Table~\ref{tab:crossmodel}, Section \ref{sec:crossmodel}).

Configs~C (model-generated replies) and D (template-based replies) yield near-identical tool selection performance as shown in Table~\ref{tab:crossmodel}. Analyzing tool call logs, we find that  Config~D reduces response length by 44\% and increases byte-identical query repeatability for Qwen. Overall, Config~D improves task completion over Config~C by 2.7, 7.9, and 10.0 percentage points for Qwen, Mistral, and Llama. These results suggest that templating provides the greatest benefit where models exhibit weaker summarization abilities. However, Config~D only templates the final tool call, passing hash-existence outputs, error payloads, and non-tool turns through as raw text.

\vspace{1pt}
\subsection{Design Guidelines for MCP Integration}
Our study yields five guidelines for efficient, secure tool-boundary design. First, routing parameters should be overridden from authoritative state, so that the server discards model-supplied values wherever it already knows the correct backend. This is safe only when the key selecting the stored record is itself trustworthy, so a supplied job ID should be checked against the active session. Second, identity should be stripped and injected server-side, because a model-supplied identity is accepted on the model's word alone and models fill the slot with placeholders when given no channel, making injection a security boundary rather than a convenience. Third, conversation history should be withheld from tool-grounded operations, since a history containing a successful prior turn was the only condition under which any model fabricated an outcome. Fourth, trust labels should vouch for outcomes rather than invocations, which requires an outcome tier so that an erroneous tool never earns the label, paired with a metric that flags success claims on turns with no tool call. Lastly, tool outputs should be scrubbed before the model sees them and the boundary must be tested against every deployed model. This is because tool return values ultimately influence the metadata while fallback behavior varies sharply across models.

\subsection{Limitations}\label{sec:limitations}
We discuss several limitations that support our observations and results. The 15 always-on decisions are integral to the architecture and were assessed by inspection and code review rather than ablation. We ran no configuration pairing history retention with identity injection, and we note that Config~A fixture asymmetry potentially understates the architecture's contribution. Because reference jobs never completed for Mistral, Qwen~2.5 72B, and Qwen~3 235B sweeps, the download benefit of B5 and B6 rests on Qwen and Llama evidence alone. Further, an invented job ID defeats both B4 and B5, and our experiments do not test for unsignaled failures such as stale or corrupted payloads~\cite{sethi2026fabrication}. Additionally, our fabrication finding rests on seven records from two models in one configuration. Lastly, the substring grading rubric ignores tool identity and outcome, and all queries use a single test user on open-weight models alone. {In our future work, our plan is to evaluate multi-user, multi-session, and richer authorization scenarios.}

\section{Conclusion}
Zero-knowledge verifiable computation is reliably accessible through natural language interfaces. Using skeptical tool validation and structured reply templates, our system improves task completion significantly over a naive chatbot baseline across all three model families, achieving 93\% on Qwen 3 32B and completing the full prove–status–download–verify lifecycle in every trial. Through experiments, we show that correctness must be enforced at the tool layer rather than delegated to the model. Without stored proof-type metadata, we find that default values are highly susceptible to routing status queries to the wrong prover. We find that tools that only fill missing values produce errors which reinforces the belief that using plausible defaults could be more hazardous than supplying no values. We further find that fabrication occurs only when conversation history contains a prior success to imitate, thereby motivating stateless tool design. Next, the results show that evaluating agent honesty against only the first tool call leads to systematic measurement errors. Lastly, we find that commitments, job IDs, and account identifiers reach the model provider in every configuration, requiring redaction before model ingestion in sensitive deployments. We anticipate this pattern generalizes to any setting requiring precise, verifiable LLM outputs, while a broader evaluation is an integral part of our future work.


\section*{Acknowledgment}
The authors disclose use of generative AI tools in the preparation of this manuscript limited to grammar, syntax, and fluency improvements to author-written text.
\bibliography{main}
\bibliographystyle{ieeetr}

\end{document}